\documentclass[letterpaper]{article} 
\usepackage{aaai2026}  
\usepackage{times}  
\usepackage{helvet}  
\usepackage{courier}  
\usepackage[hyphens]{url}  
\usepackage{graphicx} 
\usepackage{natbib}  
\usepackage{caption} 
\usepackage{algorithm}
\usepackage{algorithmic}
\usepackage{amsmath}
\usepackage{cases}
\usepackage{amssymb}
\usepackage{multirow}
\usepackage{newfloat}
\usepackage{listings}
\DeclareCaptionStyle{ruled}{labelfont=normalfont,labelsep=colon,strut=off} 
\floatstyle{ruled}
\newfloat{listing}{tb}{lst}{}
\floatname{listing}{Listing}
\title{Collaborate sim and real: Robot Bin Packing Learning in Real-world and Physical Engine}
\author{
    Lidi Zhang\textsuperscript{\rm 1}\equalcontrib,
    Han Wu\textsuperscript{\rm 2}\equalcontrib, 
    Liyu Zhang\textsuperscript{\rm 3},
    Ruofeng Liu\textsuperscript{\rm 4}, 
    Haotian Wang\textsuperscript{\rm 2}, 
    Chao Li\textsuperscript{\rm 3}, 
    Desheng Zhang\textsuperscript{\rm 5}, 
    Yunhuai Liu\textsuperscript{\rm 1}, 
    Tian He\textsuperscript{\rm 2}
}
\affiliations{
    \textsuperscript{\rm 1}Peking University, \textsuperscript{\rm 2}JD Logistics, \textsuperscript{\rm 3}Zhejiang University, \textsuperscript{\rm 4}Michigan State University, \textsuperscript{\rm 5}Rutgers University\\

}

\usepackage{bibentry}

\begin{document}

\maketitle

\begin{abstract}
The 3D bin packing problem, with its diverse industrial applications, has garnered significant research attention in recent years. Existing approaches typically model it as a discrete and static process, while real-world applications involve continuous gravity-driven interactions. This idealized simplification leads to infeasible deployments (e.g., unstable packing) in practice. Simulations with physical engine offer an opportunity to emulate continuous gravity effects, enabling the training of reinforcement learning (RL) agents to address such limitations and improve packing stability. However, a simulation-to-reality gap persists due to dynamic variations in physical properties of real-world objects, such as various friction coefficients, elasticity, and non-uniform weight distributions. 
To bridge this gap, we propose a hybrid RL framework that collaborates with physical simulation with real-world data feedback. 
Firstly, domain randomization is applied during simulation to expose agents to a spectrum of physical parameters, enhancing their generalization capability. Secondly, the RL agent is fine-tuned with real-world deployment feedback, further reducing collapse rates. Extensive experiments demonstrate that our method achieves lower collapse rates in both simulated and real-world scenarios. Large-scale deployments in logistics systems validate the practical effectiveness, with a 35\% reduction in packing collapse compared to baseline methods.
\end{abstract}


\section{Introduction}
The 3-dimensional bin packing problem (3D-BPP) \cite{pan2023adjustable} refers to a classic NP-hard combinatorial optimization problem of placing items of different sizes into containers to maximize space utilization or minimize the number of containers. With a wide range of applications, there are various realistic constraints in practice. For example, in logistics, items must typically be packed according to their incoming orders, a requirement that can be modeled as a variant of 3D-BPP known as the online 3D-BPP. In practice, enhancing container space utilization directly translates to fewer containers being required for the same task, thereby increasing cargo throughput and reducing overall expenses. 

Online 3D-BPP has attracted growing research interest \cite{zhao2021learning}. Most of the work \cite{zhao2021learning} on 3D-BPP ideally assumes each placement of items as a discrete process with static physical constraints (e.g., the gravity center must be supported by packed items). However, in real-world applications, items cannot be accurately placed at the target destination due to manipulation errors, perception errors, and protection purposes. Thus, the placement process involves continuous gravity interactions (such as impact, rebound, shaking, and sliding), and the existing stability constraint may not work well, leading to item collapse. Internal demo tests show that the existing method suffered from 70\% container collapse in our real-world scene.
\begin{figure}
    \centering
    \includegraphics[width=1.02\linewidth]{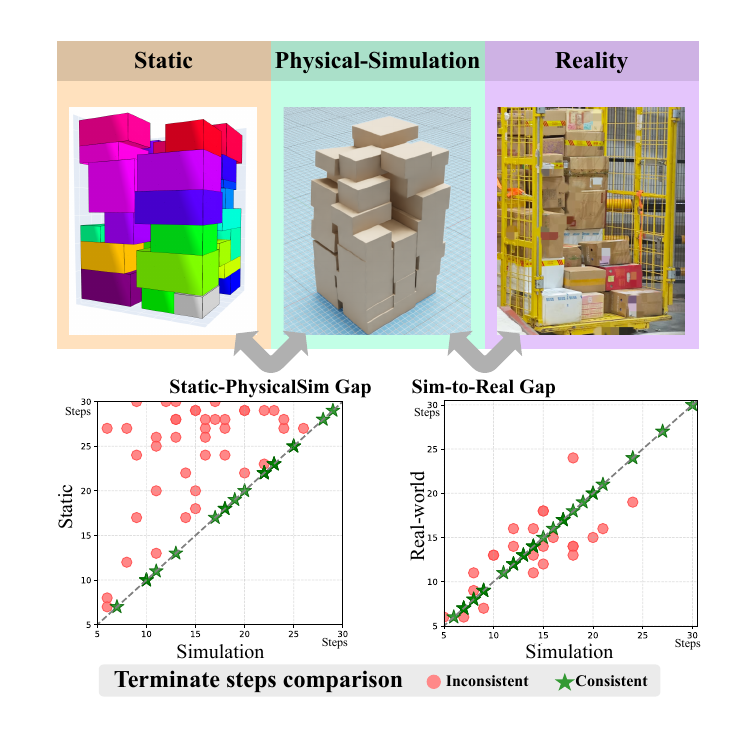}
    \caption{Gaps between different environments. The different collapse conditions lead to different termination steps across different environments. }
    \label{fig:teaser}
\end{figure}

Simulation environment with physical engine, such as Isaac Sim \cite{makoviychuk2021isaac} and Mujuco \cite{todorov2012mujoco}, offer an opportunity to solve these limitations. 
Such simulations cut time into tiny steps and calculate the interaction of objects, which can simulate and render realistic physical processes. This is also an ordinary paradigm to train an RL agent for various robotic applications \cite{gu2024humanoid}.
Although the physical engine can simulate the packing process that may fill the gap between static calculation and the physical packing process, experiments reveal that there's still a non-trivial gap between physical simulation and the real world. As shown in Figure \ref{fig:teaser}, we implement a few packing cases across three distinct environments (static calculation, physical simulation, and real world), while maintaining identical packing algorithms \cite{zhao2021learning} and input data. Under ideal conditions, this controlled setup should yield identical termination steps across all environments. Inconsistent termination steps indicate that the collapse conditions are different across different environments. 
The scatter plot demonstrates significantly greater variance between static calculation and physical simulation results, while the physical simulation shows closer alignment with real-world outcomes, though a measurable simulation-to-reality (sim-to-real) gap persists. The main reasons for this sim-to-real gap are that the weight of objects in real scenes is usually unevenly distributed, and different surface materials have different physical features, such as friction and elasticity. It is difficult and costly to obtain these parameters, thus, they cannot be consistent in simulation and reality.

To this end, we propose CoPack, a reinforcement learning (RL) framework that collaborates with physical simulation and real-world environments for robust robot packing. 
Firstly, in the RL framework, we implement domain randomization across key physical parameters in Isaac Sim, where critical physical parameters—specifically dynamic friction, static friction, and gravity distribution are systematically randomized according to manual measurements from over 100 real-world package samples.
Secondly, we collected real-world collapse cases as negative samples and fine-tuned the RL model, which can be regarded as a domain adaptation method. Then we conducted extensive experiments in a simulation environment to verify the effectiveness of CoPack. Moreover, we deployed CoPack in a real logistics center. Multi-phase experiments show that CoPack can significantly reduce the collapse of packed packages.

In summary, our contribution is threefold:
\begin{itemize}
    \item To the best of our knowledge, this work presents the first attempt to train a reinforcement learning (RL) agent for bin packing using a simulated physical engine. By incorporating physical stability constraints during training, our method significantly improves packing robustness, addressing a critical gap in real-world deployment requirements.
    \item We propose CoPack, a reinforcement learning framework that bridges simulation and reality through a hybrid training paradigm. The framework integrates: (1) physics-domain randomization (e.g., friction, mass distribution) to enhance generalization, and (2) real-world data fine-tuning for effective domain adaptation. This collaborative approach enables seamless knowledge transfer to real-world physical environments. 
    \item Extensive experiments demonstrate the superiority of our approach across both simulated and real-world benchmarks on several metrics. Notably, large-scale deployment results show a 35\% reduction in packing collapse rate compared to state-of-the-art methods, while space utility remains competitive, confirming its practical efficacy in industrial settings.
\end{itemize}
\section{Related work}
\subsection{3D bin packing}
The 3D bin packing problem (3D-BPP) has been extensively studied in offline settings, where all items are known beforehand, enabling globally optimal arrangements. Early research primarily relied on heuristic methods to find near-optimal solutions efficiently \cite{martello2000three}. More recently, deep reinforcement learning (DRL) has shown promise in surpassing \cite{zhu2021learning,hu2020tap} traditional heuristics by learning spatial reasoning from large-scale packing simulations. The online variant of 3D-BPP introduces greater complexity, as items arrive sequentially without prior knowledge of future inputs. This constraint fundamentally changes the problem dynamics, making decisions irreversible and potentially leading to cumulative inefficiencies.

Current approaches to online 3D-BPP include heuristic-based methods and DRL-based strategies. Early DRL work \cite{verma2020generalized} demonstrated generalization to arbitrary item sizes, while later advances \cite{zhao2021online,zhao2021learning} incorporated look-ahead capabilities and novel representations like configuration trees to address resolution constraints. Recent work has also explored hybrid methods \cite{yang2023heuristics}, integrating human-derived heuristics with DRL to improve real-time space utilization through synchronous packing and unpacking operations. Recently, a physics-aware method \cite{zhang2025physics} was proposed, where extra constraints are added to prevent fragile items from being damaged by heavy objects. However, these methods model the problem as a discrete process, making it impractical for real-world deployment.

\subsection{Robot packing}
Beyond theoretical space optimization in 3D-BPP algorithms, recent research has increasingly focused on incorporating practical constraints to enhance real-world applicability. For example, Wang \cite{wang2019stable} introduced a stability verification method based on force and torque balance analysis, ensuring static equilibrium for placed items. Zhao \cite{zhao2022learning} further improved the stability assessment by estimating the gravity distribution of the stacked layers, enabling the rapid evaluation of multilayered packing configurations. Furthermore, integrating computer vision techniques, such as those of Wu \cite{wu2024efficient} and Xiong \cite{xiong2023towards}, has facilitated the development of reliable robotic packing systems capable of handling end-to-end packing processes.

While these systems prioritize space utilization in lab-level small-scale deployment or simulations, real-world logistics warehouses present additional challenges. Variations in package weight distribution and material properties can significantly impact packing feasibility, rendering static computational methods inadequate in practice.

\subsection{Sim-to-real gap}

Due to their cost-effectiveness, safety, and efficiency, simulation environments such as MuJoCo \cite{todorov2012mujoco}, PyBullet \cite{benelot2018}, and Isaac Sim \cite{makoviychuk2021isaac} are widely adopted to develop and test robotic applications. These platforms greatly accelerate the iteration process of perception, decision-making, and control algorithms \cite{chevalier2023minigrid,zakka2025mujoco}.
Inherent modeling errors, including physical parameter deviations and unmodeled environmental disturbances, create a significant sim-to-real gap \cite{hofer2021sim2real} that often degrades algorithm performance during real-world deployment.
Researchers have developed several approaches to fill this gap, which can be categorized based on their reliance on real-world data. 
Domain randomization (zero-shot) \cite{horvath2022object} improves robustness through parameter randomization but may produce overly conservative policies. Domain adaptation (few-shot) \cite{xu2021learning} techniques align simulation and real-world data distributions through limited real-world samples and fine-tuning, but their effectiveness depends heavily on the availability and quality of real data. 
Data-driven methods like imitation learning \cite{belkhale2023data}, while powerful, face practical limitations due to the prohibitive costs of comprehensive real-world data collection, particularly in complex environments like warehouses.
Despite these advances, achieving reliable sim-to-real transfer for robotic bin packing remains an open challenge.


\section{Methodology}
\begin{figure}[ht]
    \centering
    \includegraphics[width=0.9\linewidth]{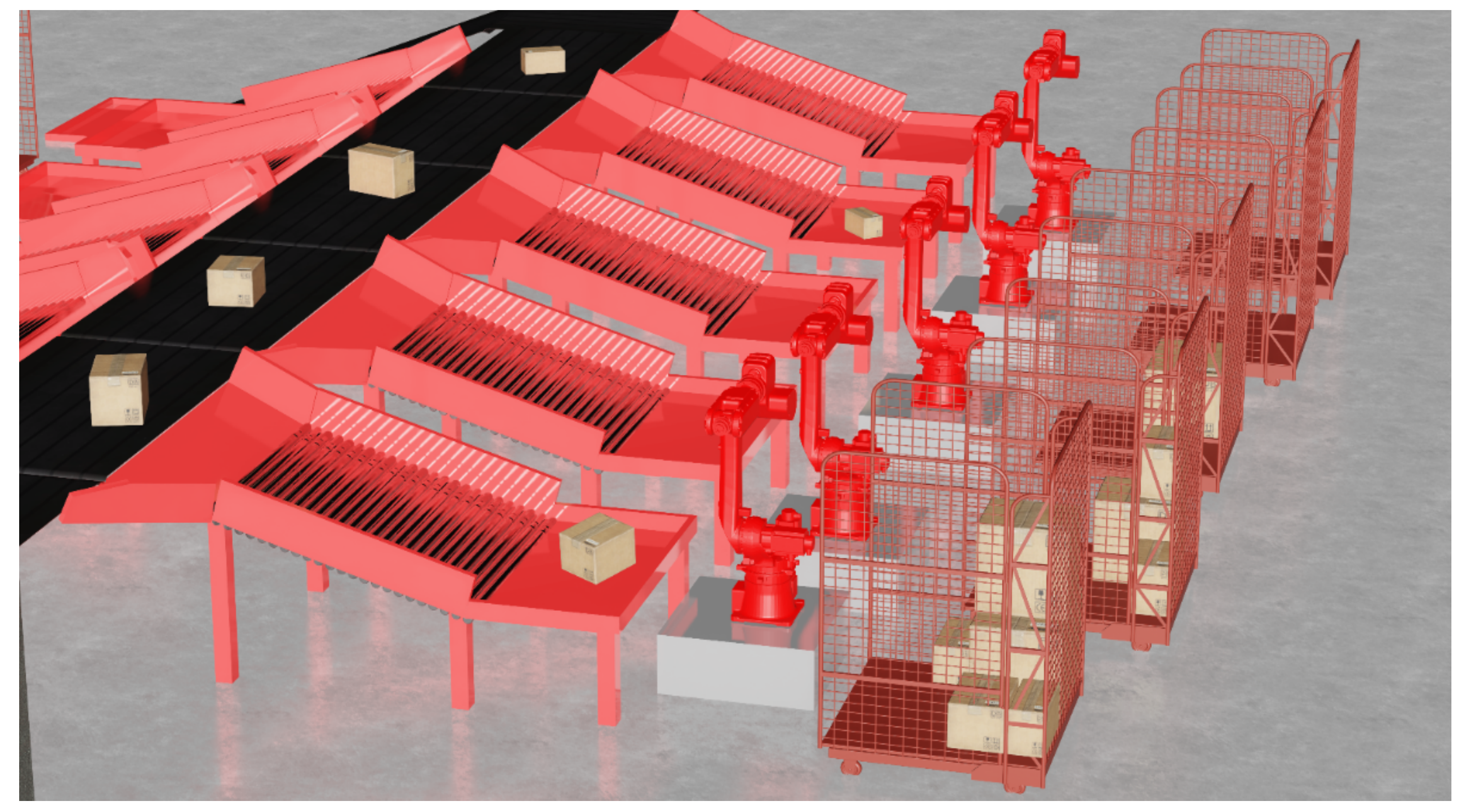}
    \caption{A sorting line in a simulated logistics center}
    \label{fig:Sim}
\end{figure}

In a logistics sorting center, a destination-driven distribution strategy is deployed to streamline item packing. As illustrated in Fig. \ref{fig:Sim}, incoming packages are automatically sorted onto designated packing platforms based on their delivery destinations. Each platform corresponds to a specific container that is pre-assigned to a particular destination. This scheme enhances the efficiency of both truck loading and unloading operations for the follow-up transportation. Based on the real-time arrival feature, the decision making of item placement position can be modeled as the following online three-dimensional packing problem.
\begin{figure*}
    \centering
    \includegraphics[width=1\linewidth]{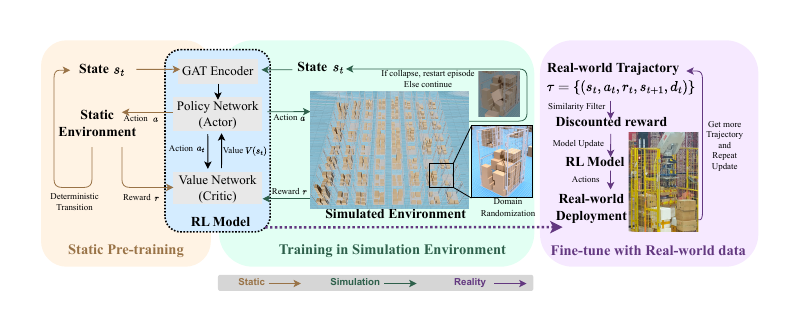}
    \vspace{-20pt}
    \caption{Overall Framework of CoPack}
    
    \label{fig:Framework}
\end{figure*}
\subsection{Problem formulation}
The online 3D bin packing problem refers to placing a set of items $i\in \mathcal{I}$ into a container with dimensions $S_x$, $S_y$, and $S_z$ in a given order.  Each item in $\mathcal{I}$ has different sizes (along x,y, and z axes) and weight, represented as $s_i^x$, $s_i^y$, $s_i^z, s_i^m$, respectively. The placement of the items is denoted by the coordinates of the front-left-bottom corner and the tilt angle, as a position set $\{(p_i^x, p_i^y, p_i^z, p_i^a)\} \in \mathcal{P}_{place}$. Generally, there is no tilt when stably packing cuboid items, that is, $p_i^a=0$. Then the online 3D-BPP can be modeled to maximize the Space Utilization (SU) as represented below.

\begin{equation} max\quad \text{SU}(\mathcal{I},\mathcal{P})=\frac{\sum_{i=0}^n{s_i^x\cdot s_i^y\cdot s_i^z}}{S_x\cdot S_y\cdot S_z}
\end{equation}
\begin{numcases}{s.t.}     
      0\leq p_i^{\text{axis}}+s_i^{\text{axis}}\leq S_{\text{axis}},\forall \text{axis} \in \{x,y,z\}, \forall i\in \mathcal{I} \\ 
      {\mathcal{V}_i }\cap{\mathcal{V}_j }=\varnothing,\forall i,j\in \mathcal{I},i\neq j
\end{numcases}
Noted that $\mathcal{V}_i$ represents the space occupied by item $i \in \mathcal{I}$. The constraints in equations (2) and (3) guarantee that all the items are placed non-overlapping and within the boundary of the container. 

Besides, there are several practical constraints in real-world deployment. 

\textbf{Stability constraint}: At each time of placement, the stability of all the packed items should be ensured. The collapse of packed items will affect the subsequent packing process, requiring manual intervention to repack, which will severely affect the efficiency of transportation. Due to friction and slight impact during placement, the packed items may have a slight position shift compared to the original placement. Therefore, we model the unstable situation as follows. For any packed item $i$ with original placement position $p_i:(p_i^x, p_i^y, p_i^z,p_i^a)$ and real-time location $p_i^t:(p_i^{x_t}, p_i^{y_t}, p_i^{z_t}, p_i^{a_t})$, it can be judged as collapsed when $d(p_i^t,p_i)$ or the tilt angle change $\Delta p_i^a$ exceeds the threshold. Noted that $d$ represents the Euclidean distance between the two coordinates.

\textbf{Feasibility constraint}:
In real-world practice, items are typically manipulated with a clamp-type or suction-type end gripper. During the packing process, it is necessary to avoid collisions between the gripper and entities such as containers and placed objects. Different picking positions directly affect the collision-free constraints during both the picking and placing phases. These coupled constraints ultimately determine the packing feasibility. For a target item $i$ characterized by its geometric occupancy $E_i$, a time-evolving container geometry $C^t$, and a suction-type gripper with geometry $E$ ($E_i,C^t,E\subset \mathbb{R}^3$). We may define $ \mathcal{P}_{pick}(i, E)$ as the set of feasible picking positions. For $p_{pick}\in \mathcal{P}_{pick}$ and $p_{place}\in \mathcal{P}_{place}$, the path planning function $f$ defined as follows:

\begin{equation}
    f(p_{pick},p_{place})_t=\begin{cases}
1 & \text{path exist, }\\
0 & \text{path does not exist.}
\end{cases}
\end{equation}
The function $f$ checks whether there is a valid path from $p_{pick}$ to $p_{place}$ at time $t$ (${t \in [0,T]}$, assuming that the whole packing process is completed within time $T$). Then a feasible packing position $p_{place}$ should satisfy:
\begin{numcases}{\exists p_{pick}\in \mathcal{P}_{pick}\quad s.t.}     
     \bigcup_{t \in [0,T]}(E^t\cup E^t_{i})\cap C^t=\varnothing \\ 
      f(p_{pick},p_{place})_t=1
\end{numcases}
where the above equations ensure that there is a valid path and no collision during movement.


\subsection{Pretraining in static environment}
Figure \ref{fig:Framework} shows the overall framework of CoPack, which consists of 3 stages: the pre-training stage in a static environment, the training stage in a physical-randomized simulation environment, and the finetune stage with real-world data as a domain adaptation. 

Since the physics engine simulation calculation speed is slow, we first use the static method to train a reinforcement learning model with the space utility as the objective function. The static method maintains almost identical training settings as the simulation engine, as described in the next part. The only difference from the physics simulation engine is that the static method uses a rule-based stability estimation method \cite{zhao2022learning} to ensure that all selected actions are theoretically feasible.

\subsection{Training scheme in physical simulation}

After the pretraining stage, we built a parallel training environment with domain-randomized physical properties for reinforcement learning in Isaac Sim \cite{makoviychuk2021isaac}, a simulation platform with a physical engine. The environment consists of multiple non-interfering containers, each executing independent packing tasks simultaneously. Our RL framework employs a Graph Attention Network (GAT) \cite{velivckovic2017graph} as the policy architecture and leverages the Actor-Critic using Kronecker-Factored Trust Region (ACKTR) \cite{wu2017scalable} method for asynchronous, iterative model updates. Each container operates as an independent training thread. If item collapse occurs within a container, all objects in that container are cleared, and a new packing episode is initialized. Within each thread, we formalize the training process as a Markov Decision Process (MDP) $\mathcal{M}=(\mathcal{S},\mathcal{A},\mathcal{T},\mathcal{R})$, defined by the following elements:

\textbf{State $\mathcal{S}$:} The current status of the container and the upcoming item to be packed. At time $t$, the state $s_t$ includes the packed items' information, location, and the current item information, denoted as $\mathcal{I}_t$, $\mathcal{P}_t$, $i_t$, respectively. 

\textbf{Action $\mathcal{A}$:} For each placement, different placement positions represent various actions. In the previous 3D bin packing algorithm, Empty maximal space (EMS) \cite{ha2017online} is considered a promising method for generating the packing action space. In this work, we additionally consider feasibility constraints and perform additional pruning of the action space.
Assuming the original action space generated with EMS is:

\begin{equation*}
    \mathcal{A}_{EMS}:=\{p_{place} | p_{place}\in \mathbb{R}^3,\text{generated with EMS}\}
\end{equation*}
Then the pruned action space under the feasibility constraint can be represented as:
\begin{equation*}
\mathcal{A}_{pru} := \mathcal{A}_{EMS}\cap \mathcal{A}_{Pick}
\end{equation*}
\begin{equation*}
\text{where } \mathcal{A}_{Pick} := \{p\in \mathbb{R}^3|\exists p' \in \mathcal{P}_{pick}, \ f(p', p) = 1\} 
\end{equation*}
The pruned actions $\mathcal{A}_{pru}$ ensure that each placement candidate has at least a valid path. In the training stage, we collect some historical data for action pruning. In the inference stage, we can obtain the feasible picking position from the upstream application.

\textbf{Transition Function $\mathcal{T}$:} Ideally, the transition $\mathcal{T}(s_{t+1}|s_t)$ is determined by the current policy. Due to the uncertainty of the real world, we randomly disturb the physical parameters in the simulation environment (friction, gravity distribution, restitution), which may cause the state transition to deviate from the ideal situation (collapse, slide, etc.). During the pre-training phase, we maintain deterministic state transitions governed exclusively by the current policy. And when training in the simulation, though the state transition becomes stochastic, we can directly obtain the current state. For the real-world deployment, we can get the container state through the camera.

\textbf{Reward Function $\mathcal{R}$:} The reward function is defined as the space utility stepwise changed after adding a new item to the container. For each successful placement of the item $i$ at time $t$, the reward is defined as intermediate reward $r_{int}(s)=s^x_i\cdot s^y_i\cdot s^z_i$. 
Packages are placed sequentially in this manner ($t<T$) until the container is filled or a collapse occurs ($t=T$). In particular, if the placement leads to collapse, the reward for the current action will be set to 0. Formally, the reward function is defined as follows:
\begin{equation}
\label{eqn:reward_function}
r(s, a) = 
    \left\{
    \begin{aligned}
        &r_t = r_{int}(s),& t<T,\  t=T\wedge f_t=1\\
        &r_t = 0,& t=T\wedge f_t=0
    \end{aligned}
    \right.
\end{equation}
where $T$ denotes the final time step, $f_t$ denotes whether the final placement is successful ($1$ for success and $0$ for failure) and $\wedge$ denotes the operator ``and".

While training the packing model in a simulation environment addresses the limitations of purely static approaches that neglect real-world physical interactions, a persistent sim-to-real gap remains due to the simulator cannot perfectly match all details of the real environment. The main sim-to-real gap in the real-world deployment of the three-dimensional packing problem is that the packages are often not perfect, rigid cuboids with uniform density. Instead, they vary slightly in shape, weight distribution, and material properties—all of which affect stability during packing. Therefore, we use domain randomization in the simulation to randomize some common physical parameters, including the friction coefficient, the center of gravity position, and placement height. For the value range of each parameter, we use a small amount of data collected in a real-world warehouse to fit the corresponding distribution. The specific parameter collection method and data details are listed in the Appendix.

\subsection{Domain adaptation: sim-and-real collaborative fine-tuning}

To further bridge the sim-to-real gap, we collect a few real-world data from the deployed prototypes and apply domain adaptation methods to fine-tune the packing model, enhancing its real-world applicability. 
To facilitate policy adaptation while maintaining behavioral consistency, we implement a trajectory-aligned fine-tuning mechanism. According to the above definition, each collected packing trajectory can be represented as:
\begin{equation}
    \tau=\{(s_t,a_t,r_t,s_{t+1},d_t)\}_{t=0}^T
\end{equation}
where $d_t$ is a binary flag representing whether the trajectory terminated due to fulfillment or collapse. 
We assign additional penalties to collapsed trajectories from the real world. Specifically, the reward for the action that leads to collapse (for $t=T\wedge f_t=0$ in Eq.\ref{eqn:reward_function}) is modified as $r_t = -\alpha r_{int}(s)$, where $\alpha$ controls the penalty strength. This penalty term assists the algorithm in more rapidly recognizing the detrimental effects of incorrect placements and promptly avoiding behaviors that lead to stacking failures. 

Given that the algorithm is continuously updated while the real-world data is generated by a previous version of the policy, there exists a distributional mismatch between the data and the current policy. During training, samples with low execution probability under the current policy but high probability under the behavior policy may be included. Since such samples have limited relevance to the current policy's improvement, we further filter them out, retaining only trajectories with high similarity to the current policy. This strategy enables faster and more efficient training. This filtered data can be modeled as:
\begin{equation}
    \{\tau|\sum_t^Td(a_t,\pi_{\theta}(s_t))<\epsilon\}
\end{equation}
where $\pi_{\theta}$ represents the policy that is currently being fine-tuned, $d(\cdot,\cdot)$ represents the distance metric and $\epsilon$ represents the threshold.

However, even after filtering, there remains a distributional bias between the retained data and the policy represented by the current samples. This mismatch can introduce errors in the estimation of the value function, leading to instability during training. To mitigate this issue, we first incorporate the importance sampling coefficient (IS) into the update equation to correct the policy update accordingly. Finally, we employ the KL divergence to constrain the distance between the updated policy and the source policy, in order to prevent excessive deviation from the original policy. The final policy Gradient equation is formally defined as:
\begin{equation}
\begin{aligned}
&\nabla_\theta J(\pi_{\theta})\\
=&\rho(s_t, a_t) \nabla_\theta log\pi (a_t|s_t)Q^\pi(s_t,a_t) - \beta KL(\pi_{\theta}|\pi_{old})     
\end{aligned}
\end{equation}
where $\rho(s_t,a_t)=\frac{\pi_{\theta}(s_t,a_t)}{\pi_{old}(s_t,a_t)}$ denotes the importance sampling coefficient, $\pi_{old}$ denotes the behavior policy of the real-world data, $\pi_\theta$ denotes the policy being updated and $\beta$ denotes the strength of the KL divergence. We stick to the previously updated paradigm for the critic updating. For more training and fine-tuning details, we have listed them in the Appendix.

\section{Experiments}
\begin{table*}[t]
\small
\centering
\begin{tabular}{lllllllllll}
\hline\hline
\multicolumn{2}{l}{\textbf{Datasets}} &
  \multicolumn{3}{c}{\textbf{Real-world Data}} &
  \multicolumn{3}{c}{\textbf{CUT1}} &
  \multicolumn{3}{c}{\textbf{CUT2}} \\ \hline
\multicolumn{2}{l}{\textbf{Method\textbackslash{}Metric}} &
  SU$\uparrow$(\%) &
  CCR$\downarrow$(\%) &
  ICR$\downarrow$(\%) &
  SU$\uparrow$(\%) &
  CCR$\downarrow$(\%) &
  ICR$\downarrow$(\%) &
  SU$\uparrow$(\%) &
  CCR$\downarrow$(\%) &
  ICR$\downarrow$(\%) \\ \hline
\multicolumn{1}{c}{\multirow{5}{*}{\rotatebox{90}{Heuristics}}} 
&
MACS &
  22.96 &
  77.8 &
  3.95 &
  44.69 &
  39.8 &
  7.01 &
  44.89 &
  45.0 &
  7.14 \\
\multicolumn{1}{c}{} &
 OnlineBPH&
  27.06 &
  74.0 &
  3.97 &
  52.12 &
  36.6 &
  6.82 &
  55.22&
  27.2 &
  6.69 \\
 &
BR &
  32.17 &
  68.6 &
  3.51 &
  43.91 &
  46.0 &
  7.31 &
  49.80 &
  27.4 &
  7.24 \\
\multicolumn{1}{c}{} &
  
  HM-Min. &
  34.58 &
  47.0 &
  4.03 &
  52.36 &
  27.2 &
  6.86 &
  52.17 &
  30.4 &
  6.99 \\
  &
  DBLF &
  50.27 &
  65.2 &
  3.88 &
  53.37 &
  37.0 &
  6.84 &
  55.65 &
  42.0 &
  7.12 \\
\hline
\multirow{4}{*}{\rotatebox{90}{RL-based}}&
 PCT&
  37.47 &
  57.6 &
  3.67 &
  58.24 &
  23.4 &
  \underline{6.43} &
  59.76 &
  12.8 &
  6.76 \\
 &
 
  LSAH&
  44.51 &
  38.8 &
  3.55 &
  54.61 &
  14.8 &
  7.29 &
  52.43 &
  31.6 &
  6.75 \\
 &
  
  GOPT &
  47.07 &
  38.6 &
  3.04 &
  52.14 &
  35.0 &
  7.05 &
  54.36 &
  13.2 &
  6.76 \\
 &
  HAC&
  51.61 &
  46.6 &
  2.88 &
  53.89 &
  12.8 &
  7.26 &
  61.68&
  12.4 &
  6.89 \\ \hline\hline \multirow{3}{*}{\rotatebox{90}{\textbf{Ours}}}
 &
  \textbf{CoPack} &
  \textbf{60.23} &
  \textbf{33.2} &
 \textbf{2.77} &
  \textbf{54.61} &
  \textbf{11.6} &
  \textbf{7.22} &
 \textbf{59.29}&
  \textbf{11.4} &
  \textbf{7.51} \\
 &
  \textbf{* w/o randomization} &
  \textbf{58.70} &
 \textbf{34.8} &
  \textbf{2.80} &
  \textbf{53.85} &
  \textbf{14.8} &
  \textbf{7.23} &
  \textbf{53.42} &
  \textbf{12.0} &
  \textbf{6.62} \\
 &
  \textbf{* w/o adaptation} &
  {\textbf{\underline{60.67}}} &
  \textbf{\underline{32.0}} &
  \textbf{\underline{2.67}} &
  \textbf{\underline{59.62}} &
  \textbf{\underline{11.4}} &
  \textbf{6.75} &
  \textbf{\underline{63.04}} &
  \textbf{\underline{8.8}} &
  \textbf{\underline{6.41}} \\ \hline\hline
\end{tabular}
\vspace{-10pt}
\caption{Comparison of CoPack and baselines. (* are the abbreviation for \textbf{CoPack}, and the \underline{underlined} are the best among all the methods.)}
\label{tab:main_result}
\end{table*}
\vspace{-10pt}

\subsection{Settings}

\textbf{Evaluation environments}: We conduct evaluations of CoPack in both simulation and real-world settings. We make all settings as close to the real world as possible, including the release height, container sizes, random gravity center bias, and random friction coefficient. We utilize Isaac Sim for simulation, as it features an integrated realistic physics engine and rendering module, which is capable of delivering high-precision physics simulation and visualization. Additionally, it is highly compatible with NVIDIA series GPUs. 

\textbf{Datasets:} We evaluated our approach using three datasets. The first two datasets (Cut1 and Cut2) are widely adopted benchmarks in three-dimensional packing research and have been consistently used in state-of-the-art studies \cite{zhao2021learning}. Moreover, we incorporated a real-world dataset comprising package records obtained from one of the largest logistics companies. This real-world dataset contains comprehensive shipment attributes, including dimensions (over 80,000 instances), weight, and arrival sequence, collected directly from operational sorting centers. See more details of datasets in Appendix. 

\textbf{Baselines:}  We utilize competitive bin packing baselines to verify the effectiveness of CoPack, including RL-based approaches (PCT \cite{zhao2021learning}, LSAH \cite{hu2017solving}, GOPT \cite{xiong2024gopt}, HAC \cite{yang2023heuristics}) and heuristic methods (DBL \cite{karabulut2004hybrid}, BR \cite{zhao2021online} HM-Min \cite{wang2019stable}, OnlineBPH \cite{ha2017online}, MACS \cite{hu2020tap}). For better adaptation of various baselines within the simulation environment, we first applied the original static methods of various baselines to solve for the packing positions of all items within each container. The item placement sequence was then input into a domain-randomized simulation environment for verification. If a collapse occurred, the episode was terminated. This protocol ensures that the stability of the packed items is positively correlated with the space utility, which is more consistent with real-world packing processes.
More implementation details are provided in the Appendix.

\textbf{Metrics:} 
We employ three key evaluation metrics: space utility (SU), container collapse rate (CCR), and item collapse rate (ICR). All the metrics are calculated as the average over 500 packing cases. The space utility follows the standard definition as previously described. Each packing process of a single container is regarded as an episode. Notably, our evaluation protocol immediately terminates the episode where collapse occurs, which inherently leads to lower space utility in dynamic environments (physical simulation and real-world testing) compared to static evaluations under identical algorithms and input data. To quantitatively assess system stability, we additionally introduce CCR and ICR. The CCR is defined as the percentage of terminated episodes due to collapse to the total number of evaluation episodes, and the ICR is defined as the percentage of item numbers that lead to collapse to the total number of items.

\subsection{Evaluation in simulation environment}
We evaluated multiple versions of CoPack against various state-of-the-art baselines on three datasets in a simulation environment. As demonstrated in Table 1, CoPack outperforms other baselines in terms of space utility and collapse rate metrics on various datasets. The performance of all baseline algorithms and CoPack algorithms on each metric on the three datasets shows different characteristics. 
CoPack maintains stable space utilization performance regardless of dataset type, while baseline algorithms exhibit marked performance degradation on real-world data.
This is because the real-world dataset has a more complex size and weight distribution, which presents greater challenges for conventional packing algorithms. The CUT series datasets exhibit significant discrepancies in ICR and CCR metrics compared to real-world datasets. This is due to the differences in item and container sizes, allowing each container in the real dataset to pack a larger number of items. However, in the Cut series dataset, each container is smaller and the item sizes have less diversity, resulting in a higher ICR than in the real datasets, while the CCR is significantly lower.

We also conducted an ablation study, comparing a CoPack model without domain adaptation and a CoPack model trained in a non-domain randomized environment. Since it is trained under a fixed physical parameter environment, the CoPack model without domain randomization performs slightly worse than the other two versions of CoPack, while the CoPack model without domain adaptation achieves better performance.  The reason is that the evaluation in Table 1 is conducted in a simulated environment; thus, different from a real-world environment, the Copack algorithm without real-world data adaptation performs better across all metrics. Combined with the actual deployment results below, the effectiveness of each component of the CoPack may be further validated.
\subsection{Large-scale deployment in real-world}

\begin{figure}
    \centering
    \includegraphics[width=0.9\linewidth]{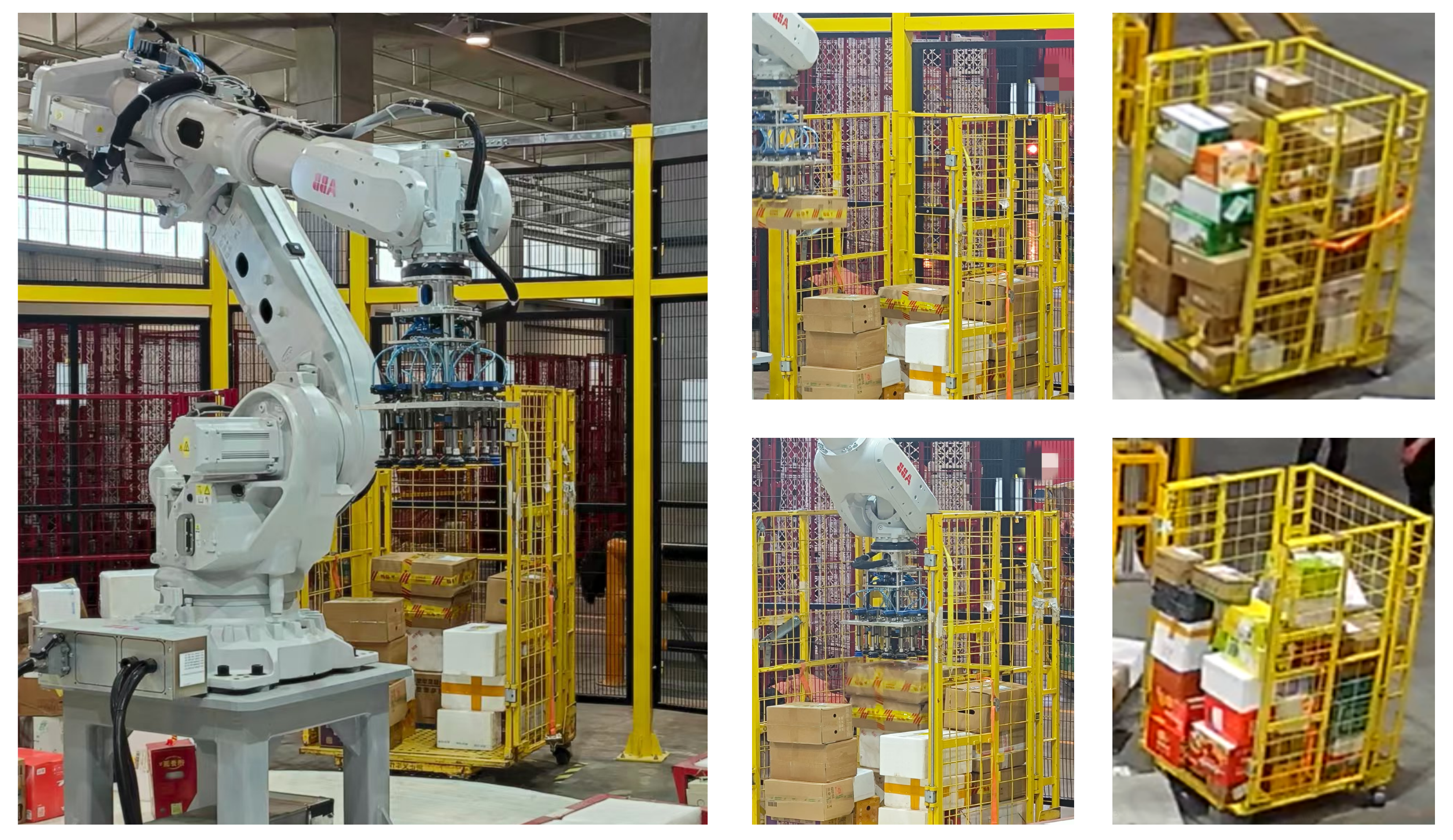}
    \vspace{-5pt}
    \caption{Deployment in logistics sorting centers with ABB Robot arm and suction-type gripper (left), packing process (middle), packed container (right). }
    \label{fig:deployrob}
\end{figure}

We deploy CoPack in several real-world logistics sorting centers in one of the biggest logistics companies globally, utilizing ABB IRB4600 robot arms equipped with custom-designed grippers integrating vacuum cup arrays, as shown in Fig. \ref{fig:deployrob}.

\begin{figure}
    \centering
    \includegraphics[width=0.9\linewidth]{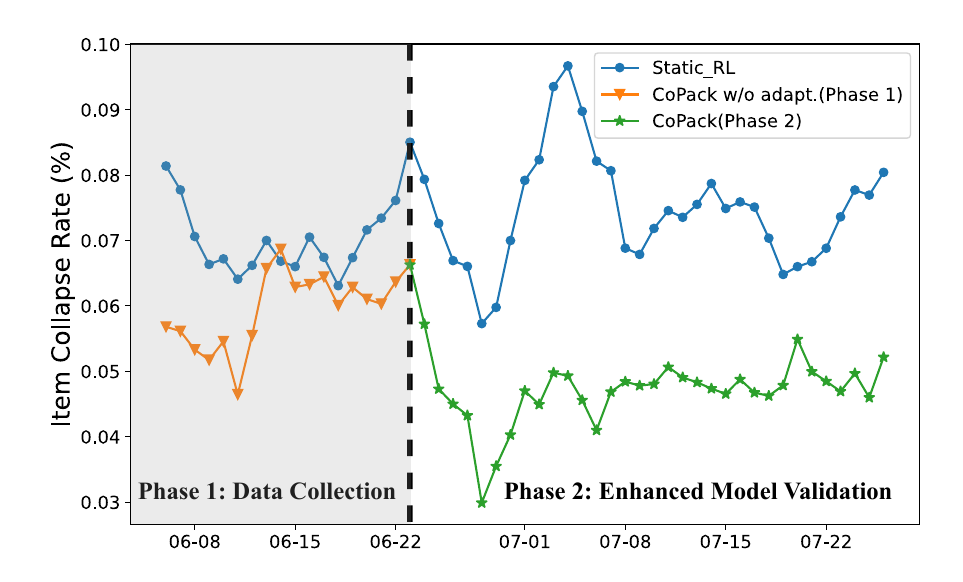}
    \vspace{-5pt}
    \caption{Item collapse rate in real-world deployment.}
    \label{fig:deploy}
\end{figure}

The CoPack framework has been incrementally deployed since June 2025, achieving cumulative processing of over 100,000 parcels across multiple facilities with dozens of robotic arms as of July 2025.
Our real-world evaluation protocol comprises two distinct phases (Fig. \ref{fig:deploy}):

\textbf{Phase 1: Data Collection} - Initial deployment of the untuned model recorded more than a thousand real collapse cases, establishing performance benchmarks.

\textbf{Phase 2: Enhanced Model Validation} - Utilizing real-world data from phase 1, the fine-tuned CoPack was implemented at the end of June, achieving lower ICR.

In practice, to maximize space utilization and handle anomalies such as collapse, our operational protocol permits human intervention to adjust the items in the container and repack the collapse recovery, after which robotic packing resumes. This operational protocol prevents us from calculating the effective container collapse rate. Therefore, we only calculate SU and ICR, where the SU metric is measured prior to any manual intervention. Although not fully eliminating human labor, this hybrid approach has substantially reduced workforce requirements at the sorting centers. We used other robotic arms equipped with a static RL method \cite{yang2023heuristics} for comparison. As demonstrated in Figure \ref{fig:deploy} and Table 2, CoPack achieves superior performance in terms of item collapse rate and space utility. In particular, the fine-tuned CoPack outperformed the original, untuned version on both metrics, further validating the effectiveness of our approach. Up to now, these robotic arms are still operating in the real world, and further collapse cases are continuously collected and used to fine-tune new versions of the CoPack model.

\begin{table}[ht]
\centering
\small
\centering
\begin{tabular}{l|ll}
Method& SU(\%)&  ICR(\%)  \\ \hline
Static RL & 46.26 & 7.31 \\
CoPack w/o adapt. (Phase 1) & 55.30 & 5.97  \\ 
\textbf{CoPack (Phase 2)}& \textbf{57.18} & \textbf{4.74} \\
\end{tabular}
\caption{Real-world demo results}
\vspace{-10pt}
\label{tab:demo_result}
\end{table}
\vspace{-5pt}
\section{Discussion}
We summarize a few key lessons learned and limitations from the implementation of CoPack.
\begin{enumerate}
    \item \textbf{Other collapse causes} - During analyzing the collapse cases one by one, we found that some of them are not caused by the improper packing position. Other collapse causes include uneven package surface, placement errors caused by motion control errors, and package drops caused by weak grip. Filling these gaps requires more sophisticated simulation or remodeling processes, which provides directions for future robot packing applications.
    \item \textbf{Upstream flow control} - In the real-world logistics center, as shown in Fig. \ref{fig:Sim}, the packing platform provides a buffer for temporarily storing several items. However, excessive item accumulation on the buffer increases collision avoidance constraints for robot grasping. Thus, in the proposed system the item flow is controlled by the upstream system to avoid item pilling on the platform.  Then, we can formulate the packing process as an online bin packing problem. This restriction can be further relaxed as the gripper capabilities improve, thereby expanding the solution space of the packing algorithm.    
    \item \textbf{Simulation training bottleneck} - We trained our model on a computer equipped with NVIDIA GTX4080 super GPU. Compared to training a packing model in a static environment ($\sim10^6$ episodes cost 200 hours), training models in a simulation environment takes more time ($\sim5\times10^4$ episodes cost 200 hours), which is the main bottleneck. Adding deformation and irregular shape features in the simulation environment will further reduce the calculation speed ($\sim10^3$ episodes cost 200 hours), making it difficult for model training to converge.

\end{enumerate}
    
\section{Conclusion}
In this paper, we propose CoPack, an RL-based robotic bin packing method, which collaborates with both physical simulation and real-world feedback. Utilizing domain randomization and adaptation, CoPack minor the gap between static packing and real-world deployment. Extensive experiments in both simulation and real-world settings show that CoPack achieves better performance on both space utility and collapse rate. In future work, we'll try to incorporate multi-modal sensory inputs (e.g., vision and tactile feedback) to develop a more robust packing system and further minimize the sim-to-real gap in robotic packing applications.

\bibliography{CameraReady/LaTeX/aaai2026.bib}

\begin{thebibliography}{29}
\providecommand{\natexlab}[1]{#1}

\bibitem[{Belkhale, Cui, and Sadigh(2023)}]{belkhale2023data}
Belkhale, S.; Cui, Y.; and Sadigh, D. 2023.
\newblock Data quality in imitation learning.
\newblock \emph{Advances in neural information processing systems}, 36: 80375--80395.

\bibitem[{Chevalier-Boisvert et~al.(2023)Chevalier-Boisvert, Dai, Towers, Perez-Vicente, Willems, Lahlou, Pal, Castro, and Terry}]{chevalier2023minigrid}
Chevalier-Boisvert, M.; Dai, B.; Towers, M.; Perez-Vicente, R.; Willems, L.; Lahlou, S.; Pal, S.; Castro, P.~S.; and Terry, J. 2023.
\newblock Minigrid \& miniworld: Modular \& customizable reinforcement learning environments for goal-oriented tasks.
\newblock \emph{Advances in Neural Information Processing Systems}, 36: 73383--73394.

\bibitem[{Ellenberger(2018--2019)}]{benelot2018}
Ellenberger, B. 2018--2019.
\newblock PyBullet Gymperium.
\newblock \url{ https://github.com/benelot/pybullet-gym}.

\bibitem[{Gu, Wang, and Chen(2024)}]{gu2024humanoid}
Gu, X.; Wang, Y.-J.; and Chen, J. 2024.
\newblock Humanoid-gym: Reinforcement learning for humanoid robot with zero-shot sim2real transfer.
\newblock \emph{arXiv preprint arXiv:2404.05695}.

\bibitem[{Ha et~al.(2017)Ha, Nguyen, Bui, and Wang}]{ha2017online}
Ha, C.~T.; Nguyen, T.~T.; Bui, L.~T.; and Wang, R. 2017.
\newblock An online packing heuristic for the three-dimensional container loading problem in dynamic environments and the physical internet.
\newblock In \emph{Applications of Evolutionary Computation: 20th European Conference, EvoApplications 2017, Amsterdam, The Netherlands, April 19-21, 2017, Proceedings, Part II 20}, 140--155. Springer.

\bibitem[{H{\"o}fer et~al.(2021)H{\"o}fer, Bekris, Handa, Gamboa, Mozifian, Golemo, Atkeson, Fox, Goldberg, Leonard et~al.}]{hofer2021sim2real}
H{\"o}fer, S.; Bekris, K.; Handa, A.; Gamboa, J.~C.; Mozifian, M.; Golemo, F.; Atkeson, C.; Fox, D.; Goldberg, K.; Leonard, J.; et~al. 2021.
\newblock Sim2real in robotics and automation: Applications and challenges.
\newblock \emph{IEEE transactions on automation science and engineering}, 18(2): 398--400.

\bibitem[{Horv{\'a}th et~al.(2022)Horv{\'a}th, Erd{\H{o}}s, Istenes, Horv{\'a}th, and F{\"o}ldi}]{horvath2022object}
Horv{\'a}th, D.; Erd{\H{o}}s, G.; Istenes, Z.; Horv{\'a}th, T.; and F{\"o}ldi, S. 2022.
\newblock Object detection using sim2real domain randomization for robotic applications.
\newblock \emph{IEEE Transactions on Robotics}, 39(2): 1225--1243.

\bibitem[{Hu et~al.(2017)Hu, Zhang, Yan, Wang, and Xu}]{hu2017solving}
Hu, H.; Zhang, X.; Yan, X.; Wang, L.; and Xu, Y. 2017.
\newblock Solving a new 3d bin packing problem with deep reinforcement learning method.
\newblock \emph{arXiv preprint arXiv:1708.05930}.

\bibitem[{Hu et~al.(2020)Hu, Xu, Chen, Gong, Zhang, and Huang}]{hu2020tap}
Hu, R.; Xu, J.; Chen, B.; Gong, M.; Zhang, H.; and Huang, H. 2020.
\newblock TAP-Net: transport-and-pack using reinforcement learning.
\newblock \emph{ACM Transactions on Graphics (TOG)}, 39(6): 1--15.

\bibitem[{Karabulut and {\.I}nceo{\u{g}}lu(2004)}]{karabulut2004hybrid}
Karabulut, K.; and {\.I}nceo{\u{g}}lu, M.~M. 2004.
\newblock A hybrid genetic algorithm for packing in 3D with deepest bottom left with fill method.
\newblock In \emph{International Conference on Advances in Information Systems}, 441--450. Springer.

\bibitem[{Makoviychuk et~al.(2021)Makoviychuk, Wawrzyniak, Guo, Lu, Storey, Macklin, Hoeller, Rudin, Allshire, Handa et~al.}]{makoviychuk2021isaac}
Makoviychuk, V.; Wawrzyniak, L.; Guo, Y.; Lu, M.; Storey, K.; Macklin, M.; Hoeller, D.; Rudin, N.; Allshire, A.; Handa, A.; et~al. 2021.
\newblock Isaac gym: High performance gpu-based physics simulation for robot learning.
\newblock \emph{arXiv preprint arXiv:2108.10470}.

\bibitem[{Martello, Pisinger, and Vigo(2000)}]{martello2000three}
Martello, S.; Pisinger, D.; and Vigo, D. 2000.
\newblock The three-dimensional bin packing problem.
\newblock \emph{Operations research}, 48(2): 256--267.

\bibitem[{Pan, Chen, and Lin(2023)}]{pan2023adjustable}
Pan, Y.; Chen, Y.; and Lin, F. 2023.
\newblock Adjustable robust reinforcement learning for online 3d bin packing.
\newblock \emph{Advances in Neural Information Processing Systems}, 36: 51926--51954.

\bibitem[{Todorov, Erez, and Tassa(2012)}]{todorov2012mujoco}
Todorov, E.; Erez, T.; and Tassa, Y. 2012.
\newblock MuJoCo: A physics engine for model-based control.
\newblock \emph{2012 IEEE/RSJ International Conference on Intelligent Robots and Systems}, 5026--5033.

\bibitem[{Veli{\v{c}}kovi{\'c} et~al.(2017)Veli{\v{c}}kovi{\'c}, Cucurull, Casanova, Romero, Lio, and Bengio}]{velivckovic2017graph}
Veli{\v{c}}kovi{\'c}, P.; Cucurull, G.; Casanova, A.; Romero, A.; Lio, P.; and Bengio, Y. 2017.
\newblock Graph attention networks.
\newblock \emph{arXiv preprint arXiv:1710.10903}.

\bibitem[{Verma et~al.(2020)Verma, Singhal, Khadilkar, Basumatary, Nayak, Singh, Kumar, and Sinha}]{verma2020generalized}
Verma, R.; Singhal, A.; Khadilkar, H.; Basumatary, A.; Nayak, S.; Singh, H.~V.; Kumar, S.; and Sinha, R. 2020.
\newblock A generalized reinforcement learning algorithm for online 3d bin-packing.
\newblock \emph{arXiv preprint arXiv:2007.00463}.

\bibitem[{Wang and Hauser(2019)}]{wang2019stable}
Wang, F.; and Hauser, K. 2019.
\newblock Stable bin packing of non-convex 3D objects with a robot manipulator.
\newblock In \emph{2019 International Conference on Robotics and Automation (ICRA)}, 8698--8704. IEEE.

\bibitem[{Wu et~al.(2017)Wu, Mansimov, Grosse, Liao, and Ba}]{wu2017scalable}
Wu, Y.; Mansimov, E.; Grosse, R.~B.; Liao, S.; and Ba, J. 2017.
\newblock Scalable trust-region method for deep reinforcement learning using kronecker-factored approximation.
\newblock \emph{Advances in neural information processing systems}, 30.

\bibitem[{Wu et~al.(2024)Wu, Li, Zhan, Liu, Liu, and Tomizuka}]{wu2024efficient}
Wu, Z.; Li, Y.; Zhan, W.; Liu, C.; Liu, Y.-H.; and Tomizuka, M. 2024.
\newblock Efficient Reinforcement Learning of Task Planners for Robotic Palletization through Iterative Action Masking Learning.
\newblock \emph{arXiv preprint arXiv:2404.04772}.

\bibitem[{Xiong et~al.(2023)Xiong, Ding, Ding, Peng, and Xu}]{xiong2023towards}
Xiong, H.; Ding, K.; Ding, W.; Peng, J.; and Xu, J. 2023.
\newblock Towards reliable robot packing system based on deep reinforcement learning.
\newblock \emph{Advanced Engineering Informatics}, 57: 102028.

\bibitem[{Xiong et~al.(2024)Xiong, Guo, Peng, Ding, Chen, Qiu, Bai, and Xu}]{xiong2024gopt}
Xiong, H.; Guo, C.; Peng, J.; Ding, K.; Chen, W.; Qiu, X.; Bai, L.; and Xu, J. 2024.
\newblock GOPT: Generalizable online 3D bin packing via transformer-based deep reinforcement learning.
\newblock \emph{IEEE Robotics and Automation Letters}.

\bibitem[{Xu et~al.(2021)Xu, Islam, Lim, and Ren}]{xu2021learning}
Xu, M.; Islam, M.; Lim, C.~M.; and Ren, H. 2021.
\newblock Learning domain adaptation with model calibration for surgical report generation in robotic surgery.
\newblock In \emph{2021 IEEE international conference on robotics and automation (ICRA)}, 12350--12356. IEEE.

\bibitem[{Yang et~al.(2023)Yang, Song, Chu, Song, Cheng, Li, and Zhang}]{yang2023heuristics}
Yang, S.; Song, S.; Chu, S.; Song, R.; Cheng, J.; Li, Y.; and Zhang, W. 2023.
\newblock Heuristics integrated deep reinforcement learning for online 3d bin packing.
\newblock \emph{IEEE Transactions on Automation Science and Engineering}, 21(1): 939--950.

\bibitem[{Zakka et~al.(2025)Zakka, Tabanpour, Liao, Haiderbhai, Holt, Luo, Allshire, Frey, Sreenath, Kahrs et~al.}]{zakka2025mujoco}
Zakka, K.; Tabanpour, B.; Liao, Q.; Haiderbhai, M.; Holt, S.; Luo, J.~Y.; Allshire, A.; Frey, E.; Sreenath, K.; Kahrs, L.~A.; et~al. 2025.
\newblock MuJoCo Playground.
\newblock \emph{arXiv preprint arXiv:2502.08844}.

\bibitem[{Zhang et~al.(2025)Zhang, Wu, Chen, Wang, Liang, Moura, Tomizuka, Ding, and Zhan}]{zhang2025physics}
Zhang, T.; Wu, Z.; Chen, Y.; Wang, Y.; Liang, B.; Moura, S.; Tomizuka, M.; Ding, M.; and Zhan, W. 2025.
\newblock Physics-Aware Robotic Palletization with Online Masking Inference.
\newblock \emph{arXiv preprint arXiv:2502.13443}.

\bibitem[{Zhao et~al.(2021)Zhao, She, Zhu, Yang, and Xu}]{zhao2021online}
Zhao, H.; She, Q.; Zhu, C.; Yang, Y.; and Xu, K. 2021.
\newblock Online 3D bin packing with constrained deep reinforcement learning.
\newblock In \emph{Proceedings of the AAAI Conference on Artificial Intelligence}, volume~35, 741--749.

\bibitem[{Zhao, Yu, and Xu(2021)}]{zhao2021learning}
Zhao, H.; Yu, Y.; and Xu, K. 2021.
\newblock Learning efficient online 3D bin packing on packing configuration trees.
\newblock In \emph{International conference on learning representations}.

\bibitem[{Zhao et~al.(2022)Zhao, Zhu, Xu, Huang, and Xu}]{zhao2022learning}
Zhao, H.; Zhu, C.; Xu, X.; Huang, H.; and Xu, K. 2022.
\newblock Learning practically feasible policies for online 3D bin packing.
\newblock \emph{Science China Information Sciences}, 65(1): 112105.

\bibitem[{Zhu et~al.(2021)Zhu, Li, Zhang, Luo, Tong, Yuan, and Zeng}]{zhu2021learning}
Zhu, Q.; Li, X.; Zhang, Z.; Luo, Z.; Tong, X.; Yuan, M.; and Zeng, J. 2021.
\newblock Learning to pack: A data-driven tree search algorithm for large-scale 3D bin packing problem.
\newblock In \emph{Proceedings of the 30th ACM International Conference on Information \& Knowledge Management}, 4393--4402.

\end{thebibliography}
\appendix
\section*{Technical Appendix}
\section*{A. Domain Randomization settings}
\begin{figure*}[ht]
    \centering
    \includegraphics[width=1\linewidth]{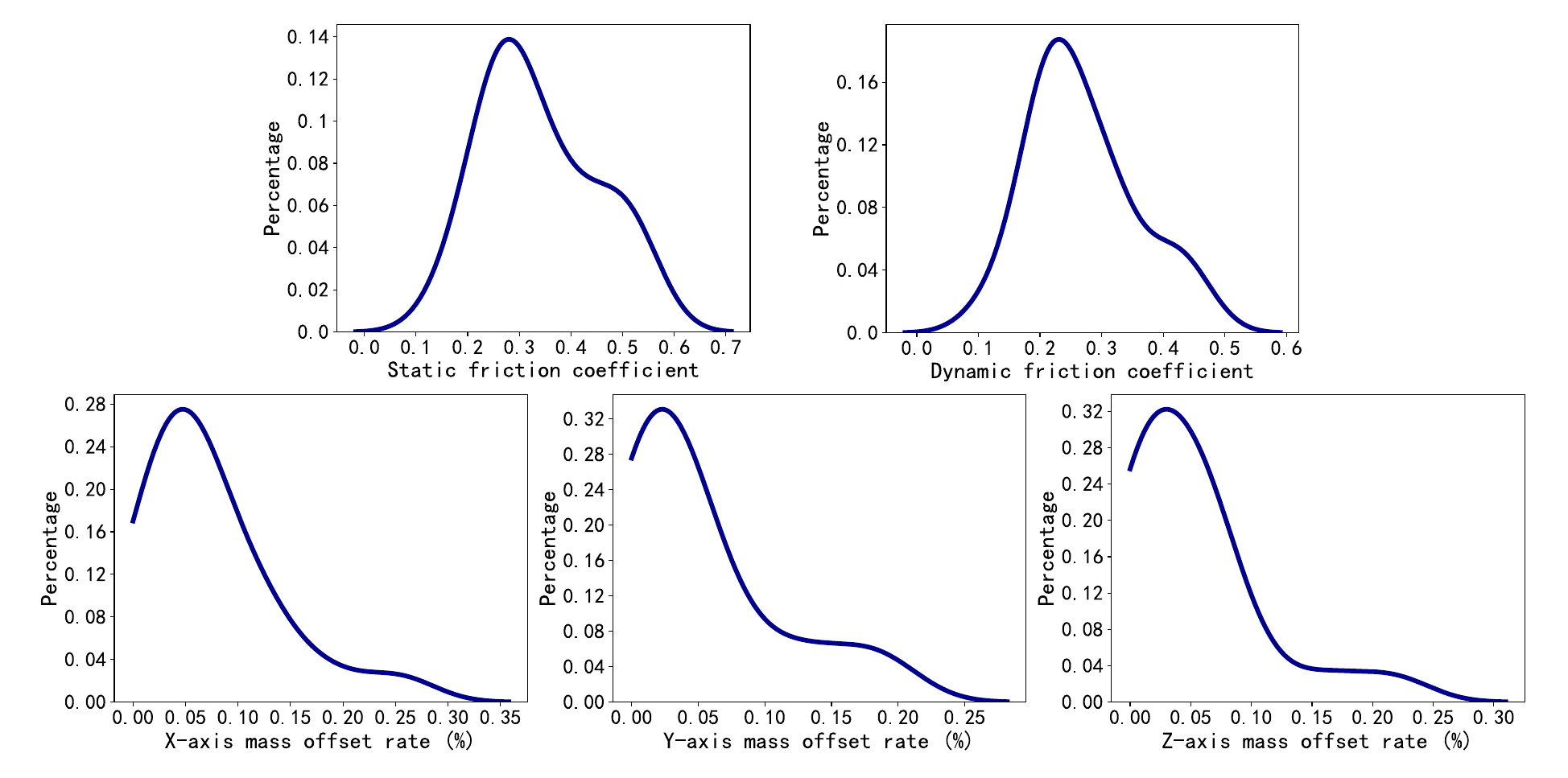}
    \caption{Fitted parameter distributions}
    \label{fig:ParaDistribution}
\end{figure*}

We randomize several physical parameters in the simulation environment, the range of the randomization is from real-world data collection. We collect data on over 100 packages in a real-world sorting center, including their static friction, dynamic friction, and mass center bias. The friction parameter was measured by dragging a small wooden block with added weights using a spring-loaded thrust meter. We first measured the maximum force required to keep the block stationary to determine static friction, and then measured the force needed to move the block at a constant speed to determine dynamic friction. The friction coefficients were calculated by dividing these force measurements by the weight of the block.
And the mass center bias was measured via a dual-suspension method. We suspended the package using two strings, each equidistant from the package’s edges, and used a spring-loaded thrust meter to measure the tension in each string. The position of the center of the mass was then calculated on the basis of the principle of levers. For each package, this measurement was repeated three times for accuracy, and the procedure was performed along all three sides of the package.
All the collected data was fitted with Gaussian Kernel Density Estimation, the fitted distribution of each parameter is shown in Figure \ref{fig:ParaDistribution}. The detailed parameter range is shown in Table \ref{friction_and_mass}. 
Moreover, we also randomize the placement height of each item, which was set to 0-5cm.

\begin{table}[h!]
\centering
\begin{tabular}{c|c|c|c}
     \hline\hline
      \textbf{Parameter/Stat}.& \textbf{Min} & \textbf{Avg} & \textbf{Max}\\
     \hline
     Dynamic friction coefficient & 0.12 & 0.27 & 0.45\\
     Static friction coefficient & 0.16 & 0.34 & 0.53\\ 
     X-axis mass offset rate (\%) & 0 & 7.23 & 25.35  \\
     Y-axis mass offset rate (\%) & 0 & 5.17 & 18.68  \\
     Z-axis mass offset rate (\%) & 0 & 5.12 & 21.77  \\ \hline\hline
\end{tabular}
\caption{Parameter ranges in domain randomization.}
\label{friction_and_mass}
\end{table}

\section*{B. Datasets Detail}
We utilize 3 datasets for the evaluation.
The real-world dataset contains more than 100,000 item information, which come from the scanning equipment in a logistics sorting center, including the dimensions and weight of each item. Due to the constraints of collision-free, the upstream distribution system filtered the dimension, shape, and weight of the items to guarantee that all the items can be successfully picked and placed. The maximum weight is set to 10 kg, and the size of each dimension is restricted to a maximum of 40 cm and a minimum of 5 cm. 

The Cut series datasets include thousands of items with dimensions ranging from 2-5 cm per side, originally packed in 10×10×10 cm containers. To better align with real-world applications, we uniformly scaled both item and container dimensions by a factor of 5, resulting in final dimensions of 10-25 cm for items and 50×50×50 cm for containers.

\section*{C. Training settings}
\subsection{Training in physical simulation}
The hyperparameters (HP) of training in physical simulation are listed in Table \ref{tab:hp_settings}. We set 16 paralleled containers in the physical simulation, and each container was trained to pack more than 50000 items. 
\begin{table}[htbp]
    \centering
    \begin{tabular}{l|c}
        \hline
        \hline
        \textbf{Hyper-parameter (HP)} & \textbf{Value} \\
        \hline
        Rollout length & 5 \\
        Paralleled episodes& 16 \\
        Batch size & 80 \\
        Leaf node holder & 80 \\
        Internal node holder & 360 \\
        Discount factor $\gamma$ & 1.0 \\
        GAT layer num & 1 \\
        Embedding size & 64 \\
        Hidden size & 128 \\
        \hline
        \hline
    \end{tabular}
    \caption{Hyper-parameter settings for simulation training.}
    \label{tab:hp_settings}
\end{table}
\subsection{Finetune settings}
We first list the hyperparameters (HP) below:

\begin{table}[htbp]
    \centering
\begin{tabular}{c|c|c|c|c}
     \hline\hline
     \textbf{HP}& \textbf{Learning rate}&\textbf{Epochs}&\textbf{Batch size}& $\alpha$\\
     \hline
     \textbf{Value}&$10^{-4}$&10&16& 0.33\\
     \hline\hline     
\end{tabular}
\caption{Hyper-parameter setting for finetune.}
\end{table}

Furthermore, we introduce the algorithm for the policy evaluation stage. The loss function of the policy evaluation stage is:

\begin{equation}
\begin{aligned}
    \mathcal{L}=&\big[r(s_t,a_t)+(1-d(s_t,a_t))Q^{target}(s_{t+1}, a_{t+1})\\
    &-Q(s_t, a_t)\big]^2
\end{aligned}
\end{equation}
where $Q^{target}(s, a)$ denotes the target value function, which updates with learning. 

However, we find it unreasonable to treat the real-world action that causes package collapse as the terminal point of an RL episode. This is because the simulation environment may assume such actions do not lead to collapse; as a result, the episode does not terminate (i.e., $d(s_t, a_t \neq 0)$), and the Q-target remains valid. This mismatch between the real world and the simulation introduces significant loss during fine-tuning and severely disturbs the Q-network’s parameter updates. Experimental results show that without proper handling, even the Q-values of normal actions are adversely affected.

To address this, we initially treat such collapse-inducing actions as normal actions and compute their Q-targets using the simulation environment. This allows real-world data to be first aligned with the simulated dynamics. We then gradually decay the target Q-value toward zero. Through this progressive adjustment, we are able to fine-tune the RL agent to align with real-world dynamics without degrading the estimation of normal actions.

\section*{D. Data Privacy and Ethical Statement}
All the presented data and figures in this paper have been strictly reviewed by the company's security department. Due to strict commercial confidentiality, we are unable to release the source code and the private datasets, which originate from a real express company. Company-reviewed codes and datasets will be made public with a license that allows free usage for research purposes after the paper is published.

\end{document}